\documentclass[letterpaper, 10 pt, conference]{ieeeconf}  
\usepackage{amsmath,amssymb,mathrsfs}
\usepackage{tikz}
\usepackage{hyperref}

\IEEEoverridecommandlockouts                              

\title{\LARGE \bf
Angular Divergent Component of Motion: A step towards planning Spatial DCM Objectives for Legged Robots
}

\author{Connor~W.~Herron$^1$,~Robert~Schuller$^2$,~Benjamin~C.~Beiter$^1$,~Robert~J.~Griffin$^3$,~Alexander~Leonessa$^1$,\\~and~Johannes~Englsberger$^2$  
\thanks{*This work was not supported by any organization}
\thanks{$^{1}$This author is with the Department of Mechanical Engineering, Virginia Polytechnic Institute and State University in Blacksburg VA, USA. (email: cwh@vt.edu, bbeiter1@vt.edu, aleoness@vt.edu)}%
\thanks{$^{2}$This author is with the Institute of Robotics and Mechatronics, German Aerospace Center (DLR), 82234 Weßling, Germany. (email: robert.schuller@dlr.de, johannes.englsberger@dlr.de)}%
\thanks{$^{3}$This author is with the Florida Institute for Human and Machine Cognition (IHMC), 40 S Alcaniz St, Pensacola, FL 32502, USA. (email: rgriffin@ihmc.org)}%
}

\begin{document}


\maketitle
\thispagestyle{empty}
\pagestyle{empty}

\begin{abstract}

In this work, the Divergent Component of Motion (DCM) method is expanded to include angular coordinates for the first time. This work introduces the idea of spatial DCM, which adds an angular objective to the existing linear DCM theory. To incorporate the angular component into the framework, a discussion is provided on extending beyond the linear motion of the Linear Inverted Pendulum model (LIPM) towards the Single Rigid Body model (SRBM) for DCM. This work presents the angular DCM theory for a 1D rotation, simplifying the SRBM rotational dynamics to a flywheel to satisfy necessary linearity constraints. The 1D angular DCM is mathematically identical to the linear DCM and defined as an angle which is ahead of the current body rotation based on the angular velocity. This theory is combined into a 3D linear and 1D angular DCM framework, with discussion on the feasibility of simultaneously achieving both sets of objectives. A simulation in MATLAB and hardware results on the TORO humanoid are presented to validate the framework's performance.

\end{abstract}

\section{Introduction}

Over the past few decades, humanoid robot locomotion remains a challenging topic due to the high number of degrees of freedom (DoF), nonlinear dynamics, and reliance on ground contact to execute stable motion. For achieving real-time implementation, full-order robot models were remapped to ``reduced-order" or template models such as the LIPM \cite{englsberger2015three, griffin2016model, hopkins2015dynamic} and SRBM \cite{bledt2020regularized, chignoli2021humanoid}. These template models were bridged with ``stability criteria" to analytically embed feasibility constraints on the external forces applied to the center of mass (CoM) by the feet to achieve walking motions \cite{vukobratovic1972on, pratt2006capture, takenaka2009real}. One of the most popular approaches of these ``stability criteria" is Divergent Component of Motion (DCM) \cite{englsberger2015three, mesesan2023unified, takenaka2009real}.

Similar to the LIPM, the DCM framework often only considers linear dynamics and conveniently handles CoM tracking by separating it into stable and unstable components \cite{englsberger2015three}. By introducing various coordinates, DCM manages to indirectly stabilize the CoM while guaranteeing the resulting external force vectors are within the base of support \cite{mesesan2023unified, griffin2016model, hopkins2014humanoid}. A useful assumption for most legged walking behaviors is that the rate of change of the centroidal angular momentum (CAM) remains small and oscillating about zero \cite{schuller2022online, seyde2018inclusion}. 
For DCM, this assumption is true if the robot's Center of Pressure (CoP) coincides with the enhanced Centroidal Momentum Pivot (eCMP) point, a position coordinate that encodes the external force vector \cite{englsberger2015three}.
However, as highlighted in \cite{schuller2022online, seyde2018inclusion}, this assumption does not hold during walking where the swing legs produce non-neglible CAM. 
While humans and humanoid robots share the capacity to swing their arms to compensate the resulting CAM generated by the legs \cite{mineshita2020jumping, khazoom2022humanoid}, certain circumstances such as locomanipulation prevent the ability to execute arm swinging behaviors. 
Research suggests that the CAM trajectory can be learned online using prior motion sequences \cite{schuller2022online} and predicted by using the known leg trajectories \cite{seyde2018inclusion} to correct the eCMP trajectory for reducing contact torques and therefore achieving CoP tracking.
The drawback of these approaches is that they are not designed for planning angular motion trajectories.
Instead, our work presents an angular component for DCM which can allow for direct rotation planning and CoP tracking. 


Alternatively, Model Predictive Control (MPC) has become a popular approach for real-time planning of quadrupedal robots often utilizing the SRBM \cite{di2018dynamic, bledt2020regularized, chignoli2021humanoid}. The SRBM includes linear and rotational dynamics and has produced highly agile results on quadrupedal robots, capable of dealing with unmodelled effects using methods such as regularization heuristics \cite{bledt2020regularized} and deep reinforcement learning \cite{pandala2022robust}. 
However, the SRBM presents coupling challenges for humanoid robots due to inertial distribution and reliance on a singular orientation coordinate  \cite{chignoli2021humanoid, chen2023integrable, ahn2021versatile}. Recently, these challenges are being overcome using methods such as Whole-Body MPC \cite{khazoom2024tailoring} and variable inertia modelling \cite{bang2024variable}. However, while MPC frameworks are simple and offer a variety of implementations, they suffer from local minima and a strong reliance on precomputed behaviors.


\begin{figure*}[!t]
\centering
\includegraphics[width=\linewidth]{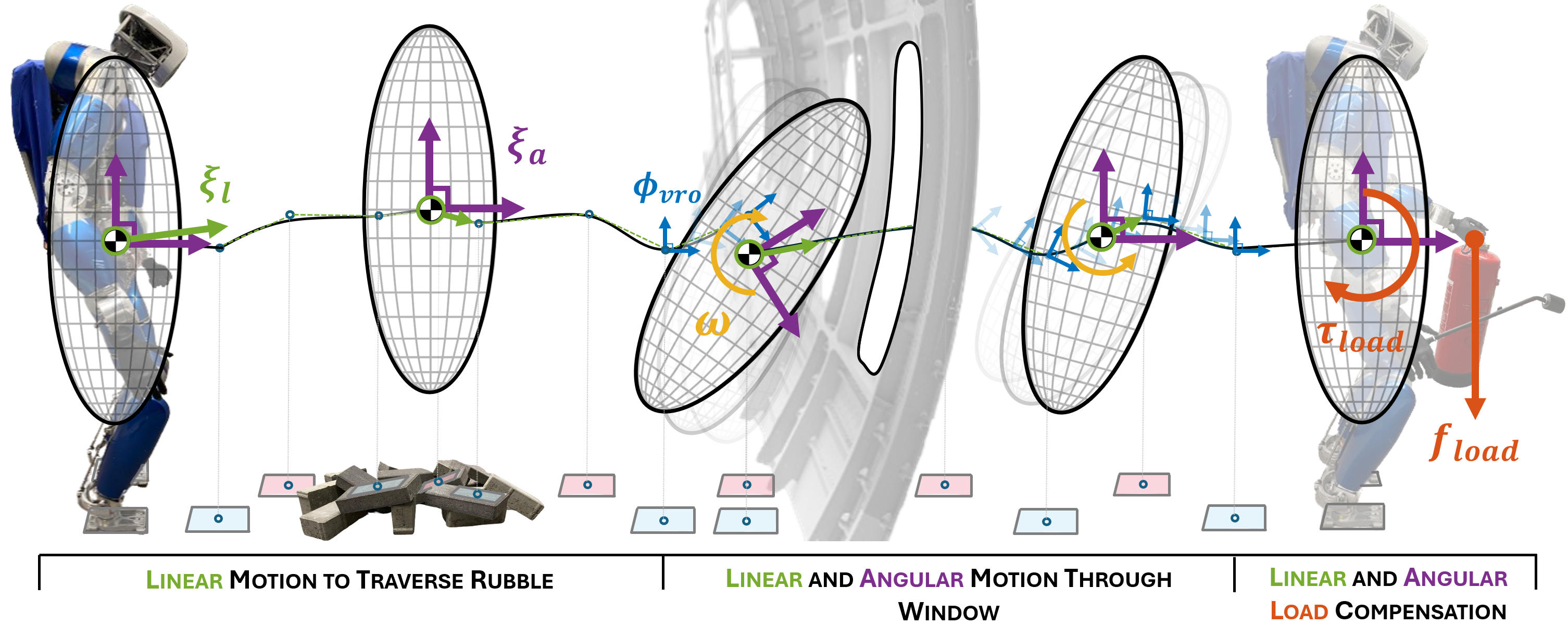} 
\vspace{-0.7 cm}
\caption{Spatial DCM is a framework allowing for linear and angular rotation planning for complex behaviors such as traversing rubble, climbing through a window, and compensating a linear and angular load.}
\label{fig:spatial-dcm-overview}
\vspace{-0.6cm}
\end{figure*}




This work focuses on extending the DCM framework to include an angular component, allowing for more direct control of the CAM for dynamic behaviors without needing any prior training.
The main contributions of this work are: 1) a description of combining linear and rotational motion into a spatial DCM framework; 2) an extension of the DCM theory to include a 1D orientation objective for legged robot planning, presented here as ``angular DCM"; 3) derivations which combine 3D linear and 1D angular DCM dynamics into a single framework; 4) a discussion on the CoP constraint and how this impacts the feasibility of simultaneously achieving linear and angular objectives; 5) MATLAB simulations and hardware experiments on the humanoid robot, TORO, which validate the performance of the proposed framework.

This paper is organized as follows: Section \ref{sec:2_spatial_dcm} discusses the idea of Spatial DCM objectives, Section \ref{sec:3_angular_dcm} describes the angular DCM theory for a 1 DoF orientation objective, Section \ref{sec:4_control_framework} combines the linear and angular DCM approaches into a compact control framework, Section \ref{sec:5_results} presents simulation and hardware results which validate the planner's performance, and Section \ref{sec:6_conclusion} is the conclusion.

\section{Spatial DCM} \label{sec:2_spatial_dcm}
As studied in several prior works such as \cite{englsberger2015three, griffin2016model}, the DCM framework often only considers the linear CoM motion to plan trajectories for legged systems while constraining the external force vectors to remain within the base of support.
As displayed in Fig. \ref{fig:spatial-dcm-overview}, this work introduces an extension of the DCM framework to include an angular component for planning spatial DCM objectives. 
For achieving spatial objectives, the SRBM is proposed where its dynamics can be defined as
\begin{align} 
    \ddot{\mathbf{x}} &= \frac{1}{m}\sum_{i=1}^n \mathbf{f}_i -\mathbf{g},\label{eqn:srb_3d_linear_dynamics} \\
    \frac{d}{dt}(\mathbf{I}\pmb{\omega}) &= \sum_{i=1}^n \mathbf{r}_i \times \mathbf{f}_i + \pmb{\tau}_i, \label{eqn:srb_3d_angular_dynamics}\\
    \dot{\mathbf{R}} &= \lfloor \pmb{\omega} \rfloor \mathbf{R}, \label{eqn:srb_3d_rotation_dynamics}
\end{align}
where $\mathbf{x} \in \mathbb{R}^3$ is the CoM position, $m$ is the mass, $\mathbf{g} = [0,\, 0,\, g]^T$ is the gravity vector, $n$ is the number of feet, $\mathbf{f}_i \in \mathbb{R}^3$ and $\pmb{\tau}_i \in \mathbb{R}^3$ are, respectively, the contact force and torque vectors of the $i$th foot, $\mathbf{r}_i \in \mathbb{R}^{3}$ is the foot position of the $i$th foot, $\mathbf{I} \in \mathbb{R}^{3 \times 3}$ is the inertia tensor in the world frame, $\pmb{\omega} \in \mathbb{R}^3$ is the angular velocity of the body, and $\mathbf{R}$ is the rotation matrix from body to world frame.

This framework would track spatial DCM objectives, $\xi = [\xi_{\rm l}^T, \, \xi_{\rm a}^T]^T$, where $\xi_{\rm l} \in \mathbb{R}^3$ is the already-known 3 DoF linear DCM \cite{englsberger2015three, griffin2016model, mesesan2023unified} and $\xi_{\rm a}$ is the angular DCM introduced in this work. 
As discussed in \cite{englsberger2015three, mesesan2023unified, hopkins2015dynamic, griffin2016model}, the linear DCM encodes velocity, acceleration, and force vectors as 3D points. Correspondingly, the angular DCM encodes rotational velocities, rotational accelerations, and torques as orientations. 
While the linear DCM is \textit{ahead} of CoM position based on the velocity vector, the angular DCM is an orientation coordinate which is \textit{ahead} of the body's rotation based on the angular velocity vector.
As displayed in Fig. \ref{fig:spatial-dcm-overview}, these motions would allow for planning complex behaviors such as traversing rubble, climbing through a narrow window, and compensating for linear and angular loads.

\section{Angular DCM Theory} \label{sec:3_angular_dcm}
\begin{figure}[t]
\centering
\includegraphics[width=0.68\linewidth]{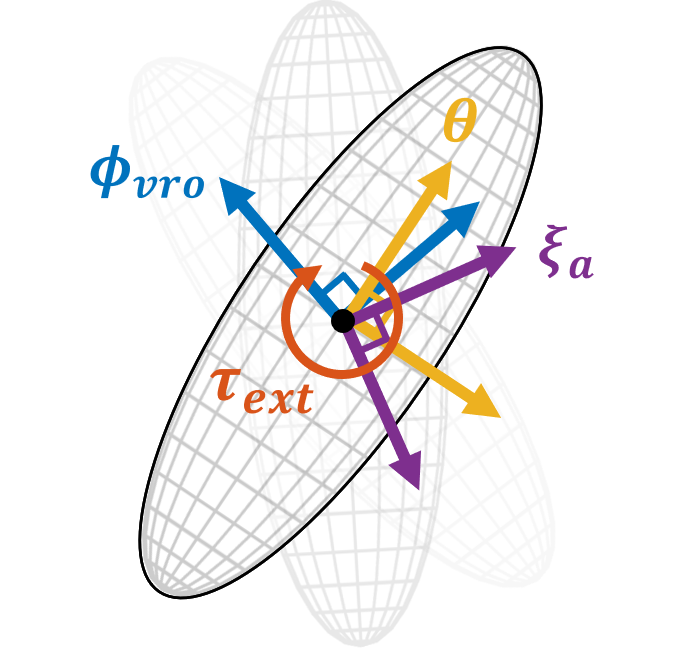}
\caption{The angular DCM, $\xi_{\rm a}$, is ahead of orientation, $\theta$, which is repelled by the VRO, $\phi_{\rm vro}$.}	
\vspace{-0.6cm}
\label{fig:srb-ang-theory}
\end{figure}

As discussed in Section \ref{sec:2_spatial_dcm}, the angular DCM, $\xi_{\rm a}$, can be defined as an orientation which is \textit{ahead} of and moving in the same direction as the body orientation. Due to challenges with nonlinearities resulting from 3D orientation dynamics, this work takes the first step by extending the DCM theory to include 1D orientation dynamics. The theory presented in this section uses a position-fixed body model, which rotates about a single axis at the centroid. In this case, the SRBM angular dynamics from (\ref{eqn:srb_3d_angular_dynamics}) and (\ref{eqn:srb_3d_rotation_dynamics}) can be simplified to
\begin{equation} \label{eqn:simple-angular-srb-dynamics}
    I\ddot{\theta} = \tau_{\rm ext},
\end{equation}
where $\tau_{\rm ext} \in \mathbb{R}$ is the torque, $I \in \mathbb{R}$ is the moment of inertia about the pivot point, and $\theta \in \mathbb{R}$ is the angular rotation of the body. 
Because the rotation dynamics are simplified and linear in (\ref{eqn:simple-angular-srb-dynamics}), the angular DCM, $\xi_a \in \mathbb{R}$, can be defined as
\begin{equation} \label{eqn:angular-dcm}
    \xi_{\rm a} = \theta + \eta \dot{\theta},
\end{equation}
where $\eta \in \mathbb{R}$ is a constant defined later on. Using (\ref{eqn:angular-dcm}), the angular DCM time derivative can be expressed as
\begin{align} \label{eqn:derivative-angular-dcm}
    \dot{\xi}_{\rm a} &= \frac{1}{\eta}(\xi_{\rm a} - \theta) + \eta \ddot{\theta}.
\end{align}
Thus, we can substitute in (\ref{eqn:simple-angular-srb-dynamics}) to find
\begin{equation} \label{eqn:derivative-angular-dcm-step2}
    \dot{\xi}_{\rm a} = \frac{1}{\eta}\xi_{\rm a} - \frac{1}{\eta}\theta + \frac{\eta}{I}\tau_{\rm ext}.
\end{equation}
At this point, we can reencode the external torque into a torsional repelling law
\begin{equation} \label{eqn:vro-law}
    \tau_{\rm ext} = \gamma(\theta - \phi_{\rm vro}),
\end{equation}
where $\phi_{\rm vro} \in \mathbb{R}$ is the Virtual Repellent Orientation (VRO) and $\gamma \in \mathbb{R}$ is a constant. The VRO encodes the external torques into an orientation that acts as a ``torsional repellant'' to rotate the body away from the VRO and towards an angular objective. By substituting (\ref{eqn:vro-law}) into (\ref{eqn:derivative-angular-dcm-step2}), we find
\begin{figure}[t]
\centering
\includegraphics[width=0.9\linewidth]{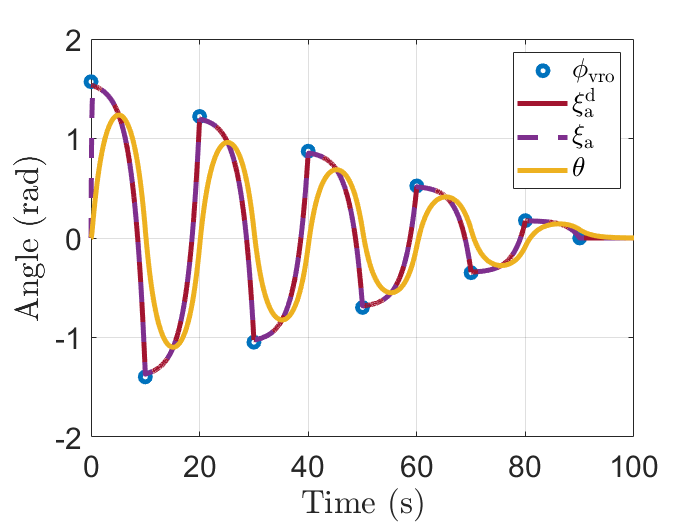}
\vspace{-0.2cm}
\caption{The VRO set-points, $\phi_{\rm vro}$, define the desired angular DCM, $\xi_{\rm a}^d$, which drives the motion of the angular DCM, $\xi_{\rm a}$, and body orientation, $\theta$.}
\label{fig:angular-dcm-sim}
\vspace{-0.6cm}
\end{figure}
\begin{equation}
    \dot{\xi}_{\rm a} = \frac{1}{\eta}\xi_{\rm a} + (\frac{\eta \gamma}{I} - \frac{1}{\eta})\theta - \frac{\eta \gamma}{I}\phi_{\rm vro}.
\end{equation}
Therefore, if we choose $\gamma = \frac{I}{\eta^2}$, we can decouple the angular body and DCM dynamics and are left with 
\begin{equation} \label{eqn:open-loop-angular-dcm-dynamics}
    \dot{\xi}_{\rm a} = \frac{1}{\eta}\xi_{\rm a} - \frac{1}{\eta} \phi_{\rm vro}.
\end{equation}
Now, we can write the decoupled dynamics in state space form as
\begin{equation} \label{eqn:open-loop-unstable-dynamics}
    \begin{bmatrix}
    \dot{\theta} \\
    \dot{\xi}_{\rm a}
    \end{bmatrix}
    =
    \begin{bmatrix}
        -\frac{1}{\eta} & \frac{1}{\eta} \\
        0 & \frac{1}{\eta}
    \end{bmatrix}
    \hspace{-0.12 cm}
    \cdot 
    \hspace{-0.12 cm}
    \begin{bmatrix}
    \theta \\
    \xi_{\rm a}
    \end{bmatrix}
    +
    \begin{bmatrix}
    0 \\
    -\frac{1}{\eta}
    \end{bmatrix}
    \phi_{\rm vro}.
\end{equation}
Notice that the open-loop angular DCM dynamics in (\ref{eqn:open-loop-unstable-dynamics}) has an unstable root. Thus, the following feedback law can be proposed for stabilizing the system and tracking a desired angular DCM trajectory
\begin{equation} \label{eqn:angular-dcm-tracking-law}
    \phi_{\rm vro} = \xi_{\rm a} + k_{\rm a}\eta (\xi_a - \xi_{\rm a}^{\rm d}) - \eta \dot{\xi}_{\rm a}^{\rm d},
\end{equation}
where $\xi_{\rm a}^{\rm d} \in \mathbb{R}$ is the desired angular DCM trajectory and $k_{\rm a} \in \mathbb{R}$ is a proportional gain. We can stabilize the closed-loop dynamics by substituting the feedback law from (\ref{eqn:angular-dcm-tracking-law}) into (\ref{eqn:open-loop-unstable-dynamics}) to get
\begin{equation} \label{eqn:closed-loop-stable-dynamics}
    \begin{bmatrix}
    \dot{\theta} \\
    \dot{\xi}_{\rm a}
    \end{bmatrix}
    =
    \begin{bmatrix}
        -\frac{1}{\eta} & \frac{1}{\eta} \\
        0 & -k_{\rm a}
    \end{bmatrix}
    \hspace{-0.12 cm}
    \cdot 
    \hspace{-0.12 cm}
    \begin{bmatrix}
    \theta \\
    \xi_{\rm a}
    \end{bmatrix}
    +
    \begin{bmatrix}
    0 & 0 \\
    k_{\rm a} & 1
    \end{bmatrix}
    \hspace{-0.12 cm}
    \cdot 
    \hspace{-0.12 cm}
    \begin{bmatrix}
    \xi_{\rm a}^{\rm d} \\
    \dot{\xi}_{\rm a}^{\rm d}
    \end{bmatrix}.
\end{equation}
In this case, $\eta$ becomes our time constant of the angular DCM dynamics occurring between the VRO setpoints. Notice that this theory is nearly identical to the definitions of linear DCM in \cite{englsberger2015three, griffin2016model,mesesan2023unified}, which is possible because of the linearity of the orientation dynamics in (\ref{eqn:simple-angular-srb-dynamics}). In Fig. \ref{fig:angular-dcm-sim}, a numerical simulation of the closed-loop tracking performance is shown from (\ref{eqn:closed-loop-stable-dynamics}). It can be seen that the body angle, $\theta$, is following the angular DCM, $\xi_{\rm a}$, while repelled by the VRO setpoints, $\phi_{\rm vro}$. This behavior is reflected in Fig. \ref{fig:srb-ang-theory}. Note that the angular DCM definition in (\ref{eqn:angular-dcm}) would need to be updated for dealing with 3D rotations because, in those cases $\frac{d}{dt}\theta \neq \omega$. 

\section{Control Framework} \label{sec:4_control_framework}
\subsection{Dynamics Derivation}
\begin{figure}[t]
\centering
\includegraphics[width=0.75\linewidth]{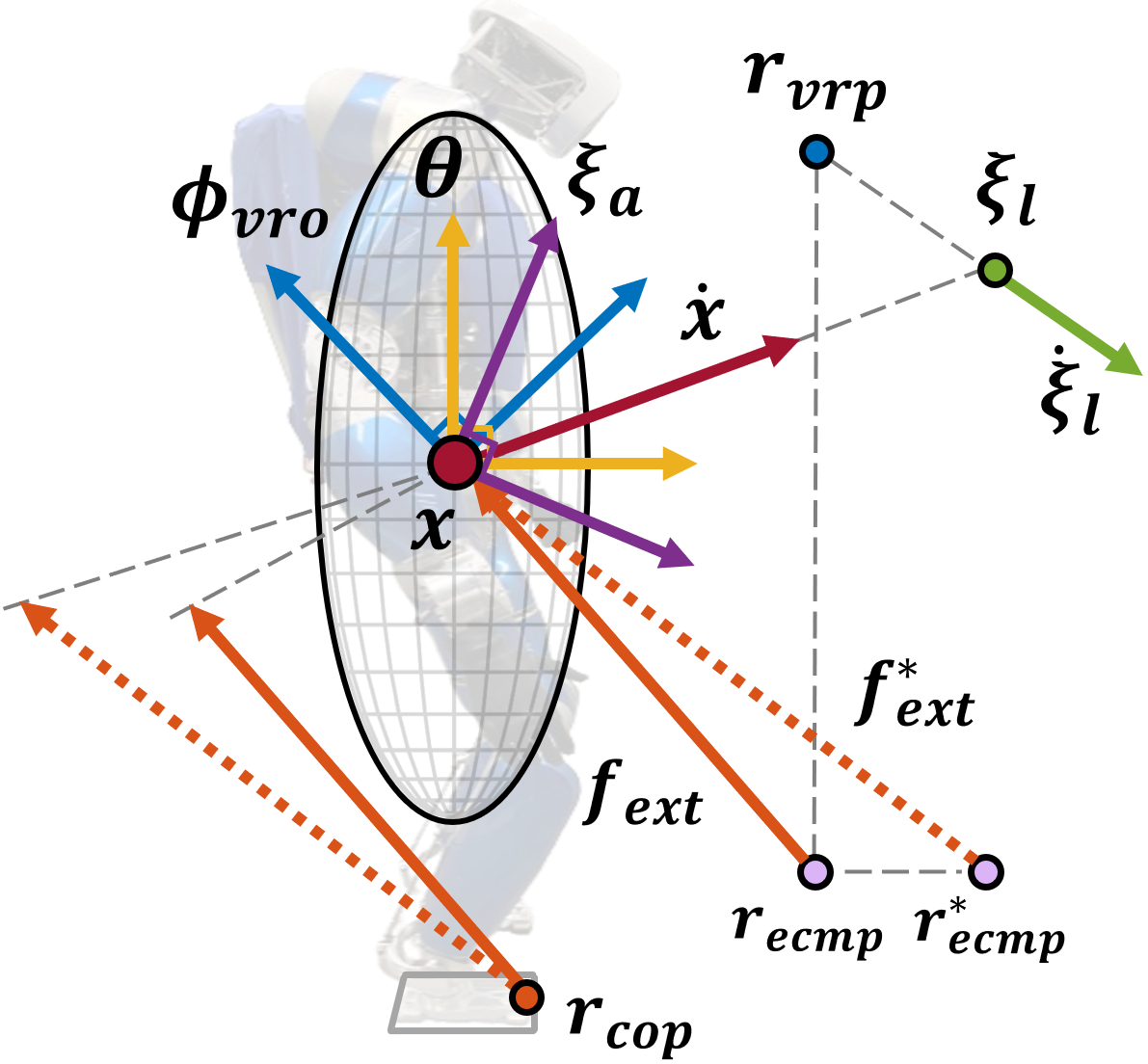}	
\vspace{-0.2cm}
\caption{Image of SRBM, angular DCM, $\xi_{\rm a}$, linear DCM, $\xi_{\rm l}$, force vector, $\mathbf{f}_{\rm ext}$, which gets projected from the CoP, $\mathbf{r}_{\rm cop}$, on the physical robot.}	
\label{fig:srb-linear-angular-dcm-layout}
\vspace{-0.6cm}
\end{figure}

This control framework combines the 3D linear and 1D angular DCM dynamics into a coherent system. In Section \ref{sec:3_angular_dcm}, the dynamics are simplified such that the body rotates about a pivot point at the centroid. For this section, the constrained SRBM can only rotate about a single axis, chosen as the pitch ($\mbox{y}$) axis, and is free to move linearly in 3D. The constrained SRBM dynamics can be simplified to
\begin{align} 
    \ddot{\mathbf{x}} &= \frac{1}{m}\mathbf{f}_{\rm ext} -\mathbf{g}, \label{eqn:srb_3d_simple_linear_dynamics} \\ 
    I\ddot{\theta} &= \mathbf{S}[\mathbf{r} \times \mathbf{f}_{\rm ext}] + \tau_{\rm ext}, \label{eqn:srb_1d_angular_dynamics}
\end{align}
where $\mathbf{f}_{\rm ext} \in \mathbb{R}^3$ is the applied external force, $\mathbf{r} \in \mathbb{R}^3$ is the footstep position in the body frame, and $\tau_{\rm ext} \in \mathbb{R}$ is the 1D external torque applied on the ground. Lastly, $\mathbf{S}$ is a selection matrix for the rotational component which matches the $\theta$ direction. For this work, $\theta$ is simply chosen as the pitch ($\mbox{y}$) angle, but in general is not restricted to a specific axis. While the cross product term seems difficult to deal with, we will show that the linear DCM dynamics are intentionally designed such that this term is cancelled. The linear DCM is defined as
\begin{align} 
    \pmb{\xi}_{\rm l} &= \mathbf{x} + b\dot{\mathbf{x}}, 
\end{align}
where $\pmb{\xi}_{\rm l} \in \mathbb{R}^3$ is the linear DCM and $\mathbf{x} \in \mathbb{R}^3$ is the CoM \cite{englsberger2015three}. The angular DCM is presented in (\ref{eqn:angular-dcm}) where $\theta \in \mathbb{R}$ is robot's base pitch angle, and  $\dot{\theta} \in \mathbb{R}$ is the robot's pitch angular velocity. We can define $b = \sqrt{h/g}$ as the DCM time constant \cite{englsberger2015three} where $h$ is the robot's height and $g$ is the gravitational constant. These DCM dynamics can be differentiated with respect to time, substituting in (\ref{eqn:srb_3d_simple_linear_dynamics}) and (\ref{eqn:srb_1d_angular_dynamics}) to get
\begin{align}
    \dot{\pmb{\xi}}_{\rm l} &= \frac{1}{b}(\pmb{\xi}_{\rm l} - \mathbf{x}) + \frac{b}{m}(\mathbf{f}_{\rm ext} -m\mathbf{g}), \\ 
    \dot{\xi}_{\rm a} &= \frac{1}{\eta}(\xi_{\rm a} - \theta) + \frac{\eta}{I} (\mathbf{S}[\mathbf{r} \times \mathbf{f}_{\rm ext}] + \tau_{\rm ext}).
\end{align}
Now, the external force, $\mathbf{f}_{\rm ext}$, is encoded using the eCMP from \cite{englsberger2015three} defined as 
\begin{equation} \label{eqn:ecmp-law}
    \mathbf{f}_{\rm ext} = s(\mathbf{x} - \mathbf{r}_{\rm ecmp}),
\end{equation}
where $s \in \mathbb{R}$ is an auxiliary constant defined later.
Therefore, the external forces and torques can be encoded using (\ref{eqn:ecmp-law}) and (\ref{eqn:vro-law}), respectively, where
\begin{align}
    \dot{\pmb{\xi}}_{\rm l} =&\,\, \frac{1}{b}(\pmb{\xi}_{\rm l} - \mathbf{x}) + \frac{b}{m}(s(\mathbf{x} - \mathbf{r}_{\rm ecmp}) -m\mathbf{g}), \\ 
    \begin{split} \label{eqn:angular-dcm-cross-term-pre-cancellation}
        \dot{\xi}_{\rm a} =&\,\, \frac{1}{\eta}(\xi_{\rm a} - \theta) + \frac{\eta}{I} (\mathbf{S}[\mathbf{r} \times s(\mathbf{x} - \mathbf{r}_{\rm ecmp})] \\
        & +  \gamma(\theta - \phi_{\rm vro})).
    \end{split}
\end{align}
The general idea of DCM is to intentionally design the external force vectors such that the CoP remains within the base of support. Therefore, based on (\ref{eqn:ecmp-law}), we assume the cross term $\mathbf{r} \times s(\mathbf{x} - \mathbf{r}_{\rm ecmp}) = 0$ since the $\mathbf{r}_{\rm ecmp}$  positions are chosen to be the footstep positions, $\mathbf{r}$. As displayed in Fig. \ref{fig:srb-linear-angular-dcm-layout} however, the external torque, $\tau_{\rm ext}$, on the physical robot is achieved via offsetting of the CoP using $\mathbf{f}_{\rm ext}$.
This assumption allows us to separate the control of linear and angular motion to the force and torque inputs, respectively, and is discussed further in Section \ref{sec:4b_cop-constraint}. Using this assumption, the linear and angular dynamics can be simplified to 
\begin{align}
    \dot{\pmb{\xi}}_{\rm l} =&\,\, \frac{1}{b}(\pmb{\xi}_{\rm l} - \mathbf{x}) + \frac{bs}{m}(\mathbf{x} - \mathbf{r}_{\rm ecmp}) - b\mathbf{g}, \\ 
    \dot{\xi}_{\rm a} =&\,\, \frac{1}{\eta}(\xi_{\rm a} - \theta) + \frac{\eta\gamma}{I} (\theta - \phi_{\rm vro}). \label{eqn:angular-dcm-cross-term-post-cancellation}
\end{align}
Finally, we can decouple the linear and angular dynamics from their DCM components by defining $s = \frac{m}{b^2}$ and $\gamma = \frac{I}{\eta^2}$ to get
\begin{align}
    \dot{\pmb{\xi}}_{\rm l} =&\,\, \frac{1}{b}\pmb{\xi}_{\rm l} - \frac{1}{b}\mathbf{r}_{\rm ecmp} - b\mathbf{g}, \\ 
    \dot{\xi}_{\rm a} =&\,\, \frac{1}{\eta}\xi_{\rm a} - \frac{1}{\eta}\phi_{\rm vro}.
\end{align}
We can also simplify the dynamics further using the Virtual Repellent Point (VRP) from \cite{englsberger2015three}, defined as 
\begin{equation}
    \mathbf{r}_{\rm vrp} = \mathbf{r}_{\rm ecmp} + b^2 \mathbf{g}.
\end{equation}
The open-loop DCM dynamics can be written in state space form as
\begin{equation} \label{eqn:open-loop-full-dynamics}
    \begin{bmatrix}
    \dot{\pmb{\xi}}_{\rm l} \\
    \dot{\xi}_{\rm a}
    \end{bmatrix}
    \hspace{-0.15 cm}
    = 
    \hspace{-0.15 cm}
    \begin{bmatrix}
        \frac{1}{b} & 0 \\
        0 & \frac{1}{\eta}
    \end{bmatrix}
    \hspace{-0.05 cm}
    \cdot
    \hspace{-0.05 cm}
    \begin{bmatrix}
    \pmb{\xi}_{\rm l} \\
    \xi_{\rm a}
    \end{bmatrix}
    \hspace{-0.05 cm}
    + 
    \hspace{-0.05 cm}
    \begin{bmatrix}
        -\frac{1}{b}\cdot \mathbf{I}_{3} & \mathbf{0}  \\
        \mathbf{0} & -\frac{1}{\eta} 
    \end{bmatrix}
    \hspace{-0.05 cm}
    \cdot
    \hspace{-0.05 cm}
    \begin{bmatrix}
        \mathbf{r}_{\rm vrp} \\
        \phi_{\rm vro}
    \end{bmatrix},
\end{equation}
where $\mathbf{I}_3$ is the $3 \times 3$ identity matrix. For stabilizing the open-loop dynamics, the linear and angular DCM dynamics require stabilizing feedback control laws. The linear DCM tracking control law presented in \cite{englsberger2015three} is of the following form
\begin{equation} \label{eqn:linear-dcm-tracking-law}
    \mathbf{r}_{\rm vrp} = \pmb{\xi}_{\rm l} + k_{\rm l}b(\pmb{\xi}_{\rm l} - \pmb{\xi}_{\rm l}^d) - b\dot{\pmb{\xi}}_{\rm l}^d,
\end{equation}
where $k_{\rm l} \in \mathbb{R}$ is the linear tuning constant and must be $k_{\rm l} > 0$ for stability purposes.
In addition, the angular DCM tracking law was previously introduced in (\ref{eqn:angular-dcm-tracking-law}). Using (\ref{eqn:linear-dcm-tracking-law}) and (\ref{eqn:angular-dcm-tracking-law}), the stable closed loop can be written as
\begin{align} \label{eqn:closed-loop-full-dynamics}
    \begin{split}
        \begin{bmatrix}
            \dot{\pmb{\xi}}_{\rm l} \\
            \dot{\xi}_{\rm a}
        \end{bmatrix}
        &=
        \begin{bmatrix}
            -k_{\rm l} & 0 \\
            0 & -k_{\rm a}
        \end{bmatrix}
        \hspace{-0.12 cm}
        \cdot 
        \hspace{-0.12 cm}
        \begin{bmatrix}
            \pmb{\xi}_{\rm l} \\
            \xi_{\rm a}
        \end{bmatrix}
         +  
        \begin{bmatrix}
            k_{\rm l}\cdot \mathbf{I}_{3} & \mathbf{I}_3 & \mathbf{0} & \mathbf{0}  \\
            \mathbf{0} & \mathbf{0} & k_{\rm a} & 1 
        \end{bmatrix}
        \hspace{-0.12 cm}
        \cdot
        \hspace{-0.12 cm}        
        \begin{bmatrix}
            \pmb{\xi}_{\rm l}^d \\
            \dot{\pmb{\xi}}_{\rm l}^d \\
            \xi_{\rm a}^d \\
            \dot{\xi}_{\rm a}^d
            \\
        \end{bmatrix}.
    \end{split}
\end{align}
The relationship between the various linear and angular DCM components can be seen in Fig. \ref{fig:srb-linear-angular-dcm-layout}. 
\begin{figure*}[!t]
\centering
\includegraphics[width=\linewidth]{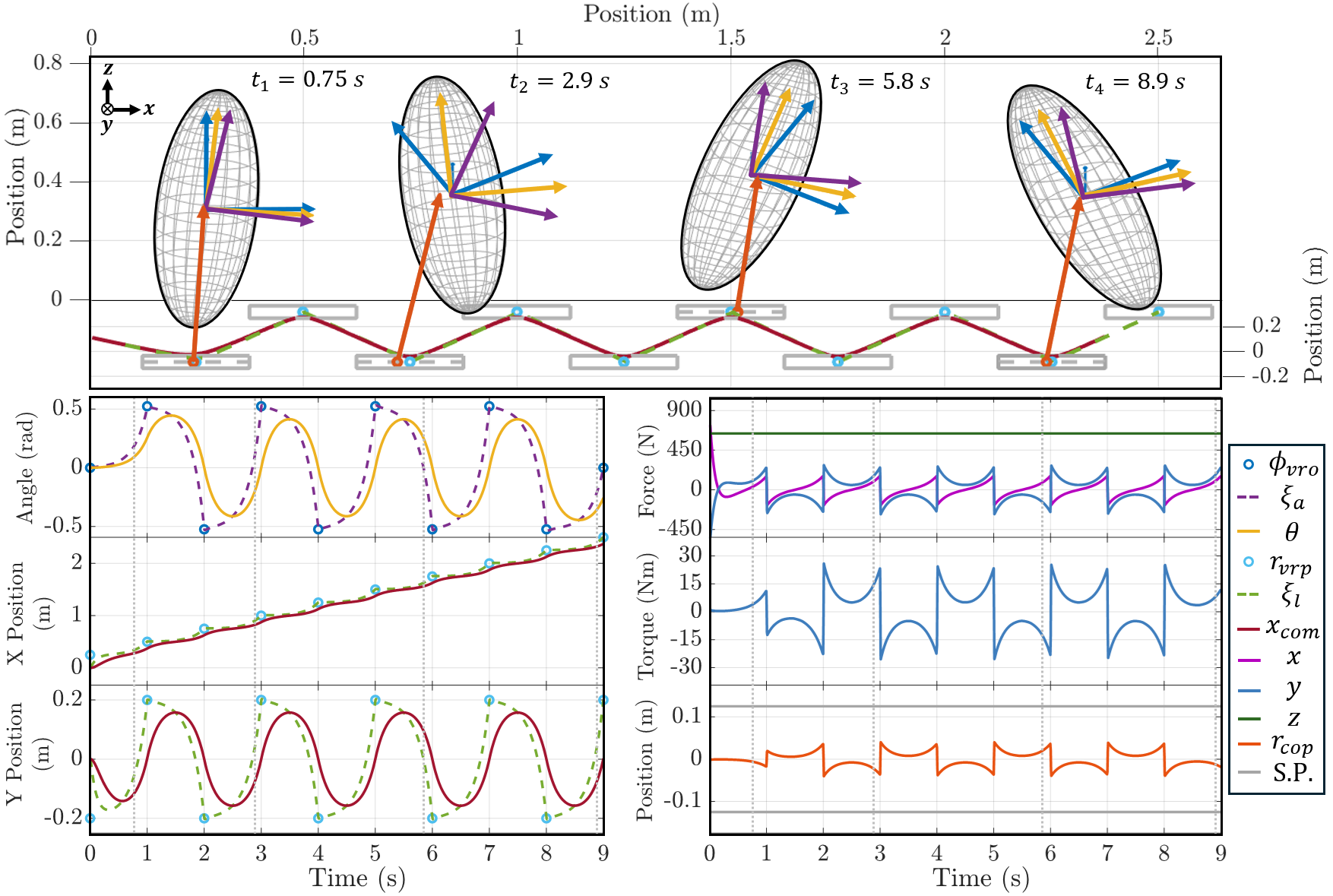}	
\vspace{-0.75 cm}
\caption{Simulation results of the 3D linear and 1D angular DCM framework simultaneously completing walking and rotating behaviors. The 3D linear motion has been projected to the ground plane.}
\label{fig:simulation-results}
\vspace{-0.6 cm}
\end{figure*}
\subsection{CoP Constraint} \label{sec:4b_cop-constraint}
This design approach intentionally decouples the linear and rotational references from the external force and  torque commands, respectively. 
On the physical robot, these linear and rotational references are feasible if the CoP stays within the support polygon. 
If the CoP exceeds the support polygon boundary, the external force or torque commands are no longer both feasible and would need to be modified. So if we exceed the CoP threshold (i.e. $|\frac{\tau_{\rm ext}}{mg}| > r_{\rm cop}^{\rm thres}$), we can redefine the external torque to 
\begin{equation} \label{eqn:external-torque*}
    \tau_{\rm ext} = \tau_{\rm ext}^{\rm max} +\bar{\tau}, 
\end{equation}
where $\tau_{\rm ext}^{\rm max} \in \mathbb{R}$ is the maximum external torque that can be applied to the ground to remain within the support polygon, and $\bar{\tau} \in \mathbb{R}$ is torque which can be generated through adjustment of the external force vector. Using a slightly modified version of (14) from \cite{hopkins2015dynamic}, we can augment the desired eCMP definition to
\begin{equation} \label{eqn:ecmp*}
    \mathbf{r}_{\rm ecmp}^* = \mathbf{r}_{\rm ecmp} + \frac{1}{m g} [\bar{\tau}_y,\, -\bar{\tau}_x,\, 0 ]^T,
\end{equation}
where $\mathbf{r}_{\rm ecmp} \in \mathbb{R}^3$ is the nominal eCMP from (\ref{eqn:ecmp-law}), and $\bar{\tau}_x, \bar{\tau}_y \in \mathbb{R}$ are the remaining external torque in the $\mbox x$ and $\mbox y$ directions, respectively, meant to be compensated by modifying the external force, $\mathbf{f}_{\rm ext}^*$. Note that $\bar{\tau} = \bar{\tau}_x$ or $\bar{\tau}_y$ depending on the rotation axis. For implementation, the angular dynamics are solved in closed-form to determine the external torque trajectory. If the resulting CoP would exceed the support polygon, then the eCMP modification in (\ref{eqn:ecmp*}) can be utilized to compensate for the additional torque requirement. While this keeps the CoP within the support polygon, the linear motion may be compromised depending on the additional torque requirement. A side effect of this approach is that the angular DCM dynamics in (\ref{eqn:angular-dcm-cross-term-post-cancellation}) is no longer valid, since $\mathbf{r} \times s(\mathbf{x} - \mathbf{r}_{\rm ecmp}^*) \neq 0$.



\section{Results and Discussion} \label{sec:5_results}




For validation of the DCM framework, a simulation and hardware experiment have been performed. The simulation is of a constrained SRBM which can move in 3-DoF with a 1-D rotation about its pitch ($\mbox{y}$) axis. The footsteps are predetermined at a constant forward walking speed with a constantly changing angular objective at each footstep. The experiment has been performed on the TORO humanoid robot with a switching angular objective ($\pm \,\pi/8\,\, \rm{rad}$) about the pitch ($\mbox{y}$) axis while standing in place.

\begin{figure*}[!t]
\centering
\includegraphics[width=\linewidth]{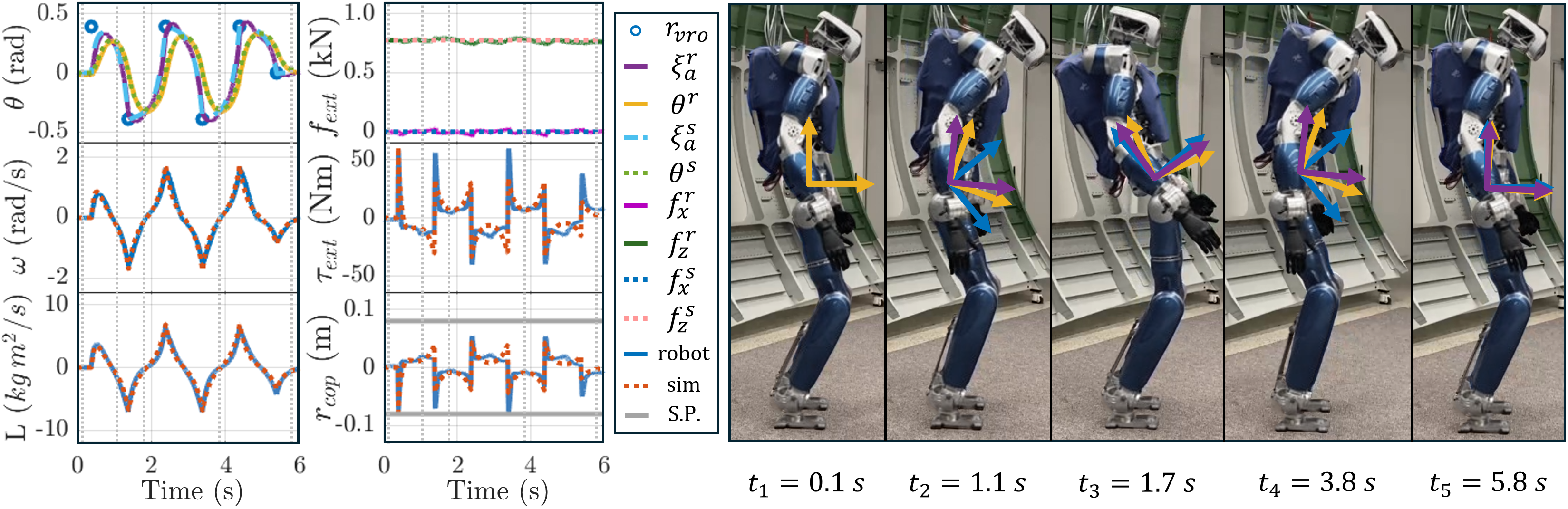}	
\vspace{-0.75 cm}
\caption{TORO accurately tracks the orientation objective from the angular DCM algorithm while standing in place.}
\vspace{-0.6 cm}
\label{fig:hardware-results}
\end{figure*}

\subsection{Simulation}
As displayed in Fig. \ref{fig:simulation-results}, a simulation has been performed in MATLAB of an SRBM which can move in 3-DoF, but is constrained to only rotate about the pitch ($\mbox{y}$) axis. The SRBM has a mass of $65.1\,\, \rm{kg}$ and an inertia about the pitch axis of $I_{\rm yy} = 2.3\,\, \rm{kg \hspace{-0.07 cm} \cdot \hspace{-0.07 cm} m^2}$, chosen to have a similar ratio to the experiment parameters. The footstep positions are predetermined based on a desired step length ($t_{\rm step} = 1\, \rm{s}$) and forward velocity ($v_x = 0.25\, \rm{m/s}$). The linear and angular DCM objectives are set for every footstep, where the linear objective is the footstep position and the angular objective is switching at $\phi_{\rm vro} = \pm\, \pi/6$. The model and controller are designed to assume an instantaneous transition between single stance phases without a double stance phase. As seen in Fig. \ref{fig:simulation-results}, the linear and angular objectives are tracked as expected and display very similar behavior despite their dynamic differences. As discussed in Section \ref{sec:4_control_framework}, the linear motion is controlled by the applied forces whereas the angular motion is controlled by the applied torque. Section \ref{sec:4b_cop-constraint} discussed how on a physical legged robot these applied wrenches are constrained such that the CoP must remain within the support polygon (denoted $\rm{S.P.}$ on the Fig. \ref{fig:simulation-results}). Considering the rotation is only about the pitch ($\mbox{y}$) axis, the applied torque and CoP position ($r_{\rm cop}$) are only 1-dimensional. The CoP (bottom right plot) achieves $4\,\, \rm{cm}$ at maximum which is far below the support polygon boundary at $\pm \,12\,\, \rm{cm}$. 

The top image displays snapshots of the SRBM during the walking with the linear and angular DCM components displayed. The angular DCM components are the VRO ($\phi_{\rm vro}$), angular DCM ($\xi_{\rm a}$), and pitch angle ($\theta$) shown as orientations at each of the snapshots as blue, purple, and gold coordinate axes, respectively. The linear DCM components shown are the VRP ($r_{\rm vrp}$), linear DCM ($\xi_{\rm l}$), and the CoM position ($x_{\rm com}$) denoted as light blue, green, and red respectively. Despite being 3D coordinates, the linear DCM components are projected to the 2D foot plane for better clarity. Finally, the GRF and CoP are denoted in orange and move along the $\mbox{x}$-axis within the foot polygon in grey. 


\subsection{Hardware}
As displayed in Fig. \ref{fig:hardware-results}, the angular DCM framework is tested on the DLR humanoid robot, TORO, whose mass is $79.2\, \rm{kg}$. Without leaving double stance, the angular DCM objective switches between $\phi_{\rm vro} = \pm\, \pi/8\,\, \rm{rad}$ every $1$ s for $5$ set points before returning to the origin. The plots on the left side directly compare the hardware response with the simulated response, denoted as $^r$ and $^s$, respectively. In addition, snapshots of TORO are featured on the right side, with the current orientations of the pitch angle (gold), angular DCM (purple), and VRO setpoint (blue). The desired orientation, angular velocity, and angular acceleration trajectories from the angular DCM framework are provided as the desired motion of the pelvis pitch orientation task for the Whole-Body Controller \cite{englsberger2020mptc}. After identifying an inertia parameter of $I_{\rm yy} = 3.96\,\, \rm{kg \, m^2}$ in post-processing, the simulation response closely resembles the hardware response on TORO. Not only does the orientation and angular velocity show excellent tracking, but the angular momentum, ground reaction wrenches, and CoP position are extremely similar between the simulation and hardware responses. Upon close inspection, small ground reaction forces in the $ \mbox x$-direction, $f_{\rm ext}^r$, are unplanned forces measured on TORO, caused by the robot nearly breaking contact due to the desired angular trajectory. The CoP trajectories reflect this behavior because they nearly cross the support polygon boundary. In addition, TORO can counteract the generated angular momentum using upper body motion \cite{schuller2021online}, but this regulation was turned off to better reflect the expected model behavior.

For expanding this work towards walking behaviors, the SRBM will encounter challenges coupling to the heavier legs of bipedal robots \cite{ahn2021versatile,chen2023integrable}. 
Future work will focus on developing a 3D angular DCM, possibly using the Whole-Body Orientation (WBO) \cite{chen2023integrable} and a variable inertia \cite{bang2024variable} for better model coupling.

\section{Conclusion} \label{sec:6_conclusion}
This work proposed the idea of spatial DCM, defining angular DCM theory for a 1D rotation, presenting a 3D linear and 1D angular DCM framework, and providing simulation and hardware results which validate the framework's performance. The angular DCM is defined to be \textit{ahead} of the orientation coordinate based on the angular velocity, identical to the linear DCM definition. The hardware performance suggests that this model and control approach can directly capture and predict the dynamics of the robot. Future work will focus on expanding the angular DCM to 3 DoF and utilizing the WBO and a variable inertia for better dynamic coupling to walking bipedal robots. 

\bibliographystyle{IEEEtran}
\bibliography{ieeeconf_backup}

\end{document}